# Robot Task Planning Based on Large Language Model Representing Knowledge with Directed Graph Structures

Yue Zhen[1)] Sheng Bi[2)] Lu Xing-tong[3)] Pan Wei-qin[4)] Shi Hai-peng[5)] Chen Zi-rui[6)] Fang Yi-shu[7)]

[1)~6)] South China University of Technology, School of Computer Science and Engineering, Guang Zhou 510006
[7)] University of Washington, Human centered design and engineering, Seattle ,98195

*Abstract* — **Traditional robot task planning methods face challenges when dealing with highly unstructured environments and complex tasks. We propose a task planning method that combines human expertise with an LLM and have designed an LLM prompt template, Think_Net_Prompt, with stronger expressive power to represent structured professional knowledge. We further propose a method to progressively decompose tasks and generate a task tree to reduce the planning volume for each task, and we have designed a strategy to decouple robot task planning. By dividing different planning entities and separating the task from the actual machine binding process, the task planning process becomes more flexible. Research results show that our method performs well in handling specified code formats, understanding the relationship between tasks and subtasks, and extracting parameters from text descriptions. However, there are also problems such as limited complexity of task logic handling, ambiguity in the quantity of parts and the precise location of assembly. Improving the precision of task description and cognitive structure can bring certain improvements.**
https://github.com/NOMIzy/Think_Net_Prompt

## I. INTRODUCTION

ROBOTS play an increasingly important role in society, and their range of applications is rapidly expanding. Traditionally, these applications were mainly concentrated in structured environments such as factories, where robot behavior is relatively fixed and often directly designated by humans. In highly unstructured human environments, such as homes, restaurants, or hospitals, robots are usually given specific goals, such as classifying and organizing specified objects, but the actions needed to achieve these goals change according to different environmental conditions. For example, to tidy up stacked bowls and chopsticks and protect them, the robot needs to pick them up in a reasonable order. In these unstructured environments, directly specifying a complete behavioral strategy for robots is impractical due to the excessive complexity of the required strategies.

The latest developments in Large Language Models (LLMs) provide a potential direction to improve the generality of robot task generation. LLMs are neural language models with a large number of parameters and trained with a large amount of data. These LLMs have shown strong universality in many natural language processing (NLP) tasks. Since the introduction of the GPT-3 model in 2020, LLMs have become an emerging research field in natural language processing and have also attracted the attention of robotics researchers.

Our goal is to combine the task semantic understanding capability of language models with the cognitive framework of humans in professional fields, to provide professional knowledge for language models, and even to train professional models, to improve their performance in professional task planning and apply it to robot task planning problems.

Current task planning work based on LLM [21,22,23,24,25] focuses on exploring the possibility and structure of generating specific content, but does not carefully consider the structure of input knowledge and further consider optimizations towards actual engineering. The work most conceptually similar to ours is [24], which considers providing a task behavior tree knowledge base to generate robot behavior trees across domains. They utilize a knowledge base storing a series of robot task behavior trees. By automatically querying the knowledge base and selecting the behavior tree most similar to the required task description as a prompt, they generate behavior trees for the required tasks in new domains. Although the focus of their work is on generating hierarchical, state machine-like structured outputs, from their work, we found the possibility of enhancing the LLM planning ability with structured knowledge. However, the tree structure that describes robot behavior similar to a state machine, such as a behavior tree, still has a lot of possiblility for improvement in terms of expressing the universal structure of knowledge for tasks in other domains. For example, a tree-thinking structure

F. A. Yue Zhen, South China University Of Technology, (e-mail: ZhenYue1614@ gmail.com/202130442396@mail.scut.edu.cn).

does not support the expression of a recursive, that is, a command with a cycle. Therefore, similar to the paradigm for describing entity relationships in the database field and the syntax paradigm in the principles of compilation, we attempt to discuss this question: Can we provide a method to better describe structured knowledge in professional fields? If progress can be made on this question, it may be possible to train general artificial intelligence in specific professional fields more efficiently. For the issue of optimization biased towards actual engineering, we have summarized our experience in the feasibility verification of Prompt and proposed some ideas for improvement and optimization.

The main contributions of this article include:

(1) Proposing an LLM prompt template, Think_Net_Prompt, which has stronger capabilities in expressing structured professional knowledge and is easy to configure, and trying to assess its feasibility. We successfully verifying the possibility of LLM using the same command to recursively layer tasks, which means that complex tasks can be analyzed in a simpler way and reduce the difficulty of professional knowledge design;

(2) Proposing a method to decompose tasks layer by layer, generating a task tree to reduce the volume of task planning each time. Proposing an executable task sequence generation algorithm which regenerates the task description and task goal according to a given precision format each time a subtask is generated, enabling LLM to perform better in single tasks .

(3) Proposing a method to decouple robot task planning at a higher level and design a method to split subtasks:

a. Divide planning entities with different professional knowledge to cooperate in generating the overall executable task sequence

b. Separate the task of binding the executable task entities according to the number of actual robots and work status and hand it over to another type of entity.

## II. BACKGROUND

*A. Robot task and action planning problem*

The problem of planning for a robot in an environment with a large number of objects, enabling it to execute actions by changing object states and its own movement in the world, is known as Task and Motion Planning (TAMP). The TAMP problem includes elements of discrete task planning, discrete continuous mathematical planning, and continuous motion planning [1]. In the most common series of solutions, the task planner uses symbolic domain statements and goals to generate candidate plan frameworks, while the action planner verifies the geometric feasibility of the plan framework and returns an action plan when successful. This process is repeated until a solution is found or the planner times out.

General TAMP methods have three core components:
1) Pre-set dynamic models and overall task state information.
2) Carefully defined symbolic planning domains, which can be adjusted for the capabilities, environment, and tasks of specific robots to perform task planning.
3) A process for testing the geometric feasibility of task plans.

Hierarchical methods [2] are characteristic of the most common solutions. They typically use:
1) AI task planners to derive candidate plan frameworks
2) Action planners to obtain action trajectories that satisfy robot and environmental constraints; for example, through sample-based planning [3] or constraint optimization [4].

Current general ideas to accelerate TAMP include: learning sampling distributions [5], visual feasibility heuristics [6,7,8], low-level controllers [9,10], or state sparsifiers [11,12]. However, these methods learn solutions computed by classic TAMP solvers, so they also rely on carefully designed symbolic planning domains specific to the task. While methods have been proposed for learning symbolic representations for TAMP [13,14], these methods usually require symbolic transformation prior knowledge specific to the task.

*B. Application of language conditioned policy in robot task and action planning problem*

Language, as a medium for solving TAMP, has attracted a lot of attention. Language conditioned policies (LCP) can now be applied to manipulate robots. Many methods have been proposed for short-term tasks [15,16,17], and some focus on long-term tasks [18,19].

*C. Related work on task planning with LLM*

The emergence of Large Language Models (LLM) as a task-independent reasoning module provides a promising path to achieve universal robot planning capabilities. Large language models can utilize a wealth of knowledge learned from a large amount of text, but they may not necessarily be able to decompose high-level commands into low-level instructions suitable for robot execution. To make the language model adapt to the problem statement and give the expected output, it needs to decompose high-level commands into a sequence of usable low-level skills.

Several recent works utilize the generative features of LLM by prompting them to generate long-term plans: [20] confines the LLM planner to a feasible set of actions, exploring the potential of language models applied to TAMP problems. Related work translates plans generated by LLM from natural language into code [21]. Utilizing LLM's ability to perform robot system planning without manually specifying the symbolic planning domain, the SayCan framework [22] combines human high-level instructions and their corresponding robot basic tasks into prompts, ProgPrompt [23] represents robot tasks as Pythonic programs, and then uses Pythonic code as prompts. Paper [24] uses a large language model to generate a three-layer behavior tree for robot task planning, demonstrating the feasibility of LLM generating structured content.

Paper [25] proposed Text2Motion, based on previous works,,

which connects LLM with a set of learned skill policy libraries and policy sequence optimizers [26] to solve geometrically complex continuous manipulation tasks, providing a promising language-based planning framework for solving continuous manipulation tasks with geometric dependencies.

The above works have made some progress in lower-level geometric dependent task planning and preliminary use of language to invoke robot commands, but at a higher level of task planning, although there have been attempts to provide LLM with more accurate, structured information for task planning [24], there hasn't been serious consideration for a general method of providing more complex structured professional knowledge for the semantic understanding capabilities of LLM. At the same time, while attempting to reproduce the prompt engineering from the above papers, we noticed some problems: a. the same prompt template, tasks descriptions of the same meaning but with different levels of precision and logic, affect the quality of the results; b. the same prompt template, as the complexity of task description logic increases, the quality of the results decreases and more errors occur.

Therefore, we propose a method that uses a directed graph structure to precisely describe the instruction set, break down tasks, and use the semantic analysis capabilities of LLM for task planning, thereby informing LLM of more professional knowledge methods that humans have, and requiring LLM to make precise statements about ambiguous task descriptions during the iterative generation process, limiting the complexity of single task planning logic in a way that allows LLM to output sequential operation sequence codes that robots can parse and execute with a high probability.

## III. RESEARCH METHODS

Considering the general method of providing LLM's semantic understanding capabilities with more complex structured professional knowledge, based on experiments on language model characteristics, we found that when planning complex tasks, if the language model is provided with some possible sub-task sequences as a reference for thinking, the model can output more likely executable sequences. These tasks, according to their possible set of sub-tasks and possible sub-task sequences, logically form an interconnected relationship in a directed graph.

Possible process optimization problems we discovered:

1) the same prompt template, tasks descriptions of the same meaning but with different levels of precision and logic, affect the quality of the results;

2) the same prompt template, as the complexity of task description logic increases, the quality of the results decreases and more errors occur.

For the first problem: The current general language model, in tasks related to numbers and spatial relations, if it is implicitly given a task description and instruction to generate a solution sequence, for example, telling it how to assemble a screw on the chassis, and then telling it that there are 7 more screws and the method is the same, install them on the chassis, such a description often results in a chaotic result. However, if it is provided with parameters to think about to make the vague description precise, such as generating the installation process description for the remaining 7 screws for this type of description, letting it re-describe the task in a form close to a function (a natural language sentence containing verbs and related parameters) as input, the quality of the generated result can be improved.

For the second problem: we reduce the complexity of single-time planning by decomposing tasks, and there are two approaches to this: one is layering. There are two possible implementations of this approach. One is to enforce layering, by directly designing task words that unfold in layers. The other is to describe the termination condition of this layered planning task in the prompt, and include this task itself among its optional subtasks, allowing the LLM to recursively layer based on its understanding. Theoretically speaking, the second method has better generalization performance. Another approach is to decouple and separate different tasks, such as separating the task of allocating specific machines. This way, the LLM does not need to consider too many problems at once, which not only improves the quality of the results, but also provides convenience for code implementation and maintenance.

### A. Professional Knowledge Mapping Mechanism

In specific fields, people conceptualize and abstract their perceptual understanding of things into rational knowledge through long-term practice. This knowledge, refined by applying it in practice and revising it according to feedback, which can correctly solve problems, is called professional knowledge. Professional knowledge includes but is not limited to concepts, relationships and laws between concepts, paradigms for thinking about specific types of problems, and best practices for solving them. For example, consider an excellent human hunter, to successfully complete a hunt, he first trains to accumulate basic operations such as fast walking, jogging, sprinting, changing direction, throwing spears, etc. Then he masters the habits of different prey and terrain features, refines reasonable hunting concepts, and organizes his basic operation process according to different task situations, verifying the reasonable and effective sequence in practice. At present, although LLM has shown powerful semantic understanding capabilities, due to the limitations of training data, it does not directly understand the professional knowledge that has been accumulated in specific professional fields. If there is a method to efficiently establish precise mappings for professional field concepts, laws, problem classifications, and general practice methods under different problems and scenarios, it may effectively improve the ability to use the general model to solve problems and plan tasks.

During the exploration process, we abstracted a generic thinking framework for planning an object assembly task sequence based on assembly guides. According to the available subtasks and possible subtask sequences when large tasks are broken down into small tasks in the thinking framework, we designed intermediate task words and designed for them a set of executable task words for a robotic arm that can complete different types of gripping and assembly steps and is equipped with visual object detection capabilities. For each task word, the content included in its data structure is as shown in the figure below, including: a flag field indicating whether this task word is terminal, the specific content mapped by this task word, the list of parameters and parameter content descriptions of this task word, possible sub-task word list, and a list of possible sub-task word sequences. Logically, these task words form a directed relationship based on two possible relationships, "this task word may generate a sub-task sequence containing another task word" and "the next task word in the task sequence where this task word exists may be this task word."

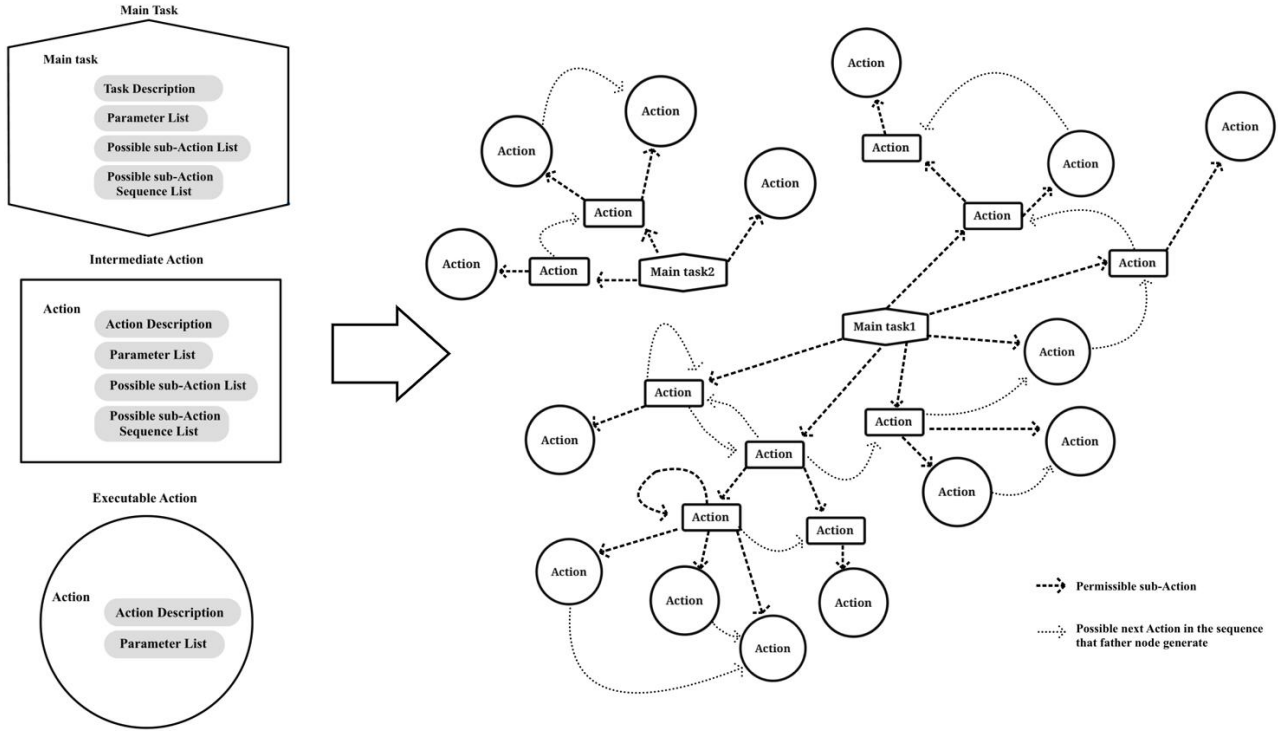

**Fig. 1.** The schematic of a directed graph constructed among different types of overall tasks, intermediate tasks requiring further generation of task sub-sequences, and executable task words that no longer generate downwards.

In order to verify the feasibility of this structure, we first use the method of the Prompt Engineering to conduct preliminary exploration. We design the input and output format in JSON, a human-readable, hierarchical, array-supported, and relatively concise coding language. This format is also easy to be parsed and processed by various high-level programming languages.

**Table. 1. (a)** Think_Net_Prompt input format.

| Input_format |
|---|
| {<br>  "task": "Action word",<br> "introduction": "brief introduction information about the task",<br> "task_parameters": {<br>  "param1":"param1_value",<br>  "param2":"param2_value"<br> },<br> "possible_subtasks": [<br>  "subtask1",<br>  "subtask2"<br> ],<br> "subtask_descriptions": [<br>  "subtask1_description",<br>  "subtask2_description"<br> ],<br> "subtask_parameters": {<br>  "subtask1": [<br>    {"name":"param1",<br>     "type":"type of this param,like int/str/float",<br>     "description":"description about this param"<br>    },<br>    {"name":"param2",<br>     "type":"type of this param,like int/str/float",<br>     "description":"description about this param"<br>    }<br>  ],<br>  "subtask2": [<br>  {"name":"param1",<br>  "type":"type of this param,like int/str/float",<br>  "description":"description about this param" |

```
            },
        {"name":"param2",
         "type":"type of this param,like int/str/float",
         "description":"description about this param"
            }
      ]
   },
   "possible_subtask_sequences": [
            ["subtask1_action","subtask2_action"],
            ["subtask2_action","subtask1_action"]
    ],
 }
```

Table. 1. (b) Think_Net_Prompt output format.

**Output_format**

```
{
   "subtask_sequence":[
      {"action":"action1",
    "parameters":{
      "param1":"param1_value",
          "param2":"param2_value"
     }
    },
       {"action":"action2",
    "parameters":{
      "param1":"param1_value",
          "param2":"param2_value"
     }
    }
  ]
}
```

### B. Executable Task Sequence Generation Algorithm

*1) Task Tree, Forest, and Generation of Executable Task Sequences*

In order to generate the final sequence of executable subtasks as output, reduce the complexity of LLM task planning in a single interaction, and support the cooperation of different entities involved in task planning during task generation, we design a step-by-step task decomposition and sequence generation method as follows: We design a tree node that can represent an instantiated task description, task word, task word instantiated parameters, which has its own subtask sequence. We use this type of tree node to organize a tree. Each branch of the tree will continuously generate sub-sequences until all the leaf nodes represent executable task words that cannot continue to generate sub-sequences. The specific process is as follows:

1. First, for a main task word, we have a root node as the starting point of the task tree.

2. During the generation of the task tree, we will use specific task words to describe the operations of each task node.

3. For each current leaf node, we will get its task word and parameters.

4. Next, we will check if this task word is in the knowledge base.

5. If the task word is valid, we will continue the following steps; otherwise, we will throw an exception.

6. In the task tree, we will continue to loop until there are no leaf nodes of executable task words that can continue to generate sub-sequences.

7. For each leaf node of an executable task word that can continue to generate sub-sequences, we will perform the following steps:

   a. Get the actions and parameters of the node.

   b. Retrieve the corresponding task word from the knowledge base.

   c. If the task word cannot be found, throw an exception.

   d. Obtain the rules and action limitations of the task word.

   e. Combine the actions, parameters, rules, action restrictions, and general information into a prompt.

   f. Send the prompt to the language model.

   g. Receive the response returned in JSON format from the language model, which includes the overall task word sequence and corresponding parameter string.

   h. Record the response log.

   i. Convert the JSON string into a dictionary.

   j. For each object in the dictionary:

     - Get the values of "task word" and "parameter list".

     - Check whether the task word exists in the knowledge base.

     - If it exists, create the corresponding task node object according to the "task word" and "parameter list" of the current object.

     - Add the newly created task node as a child node of the leaf node being processed in the current iteration.

8. Repeat steps 6 and 7 until there are no leaf nodes that are not executable task words that can continue to generate sub-sequences.

9. Complete the generation process of a task tree for a total task word.

At the very beginning, after obtaining the task description, we use similar generation logic to first generate a one-layer sequence of total task words, then generate a task tree for each total task word, thus we get a forest. By sequentially traversing and taking out all its leaf nodes, we get the executable task sequence. This design is to decouple, making the whole system more scalable, which will be specifically explained in the following text.

*2) Clarification of Ambiguous Tasks in Task Generation*

Although this action cannot be seen in the Prompt template given earlier, in fact, it can directly serve as a parameter for each task word. Descriptions of regenerate task description to more precise format can be written in the parameter description, thus achieving optimization.

*3) Cooperation in Task Generation*

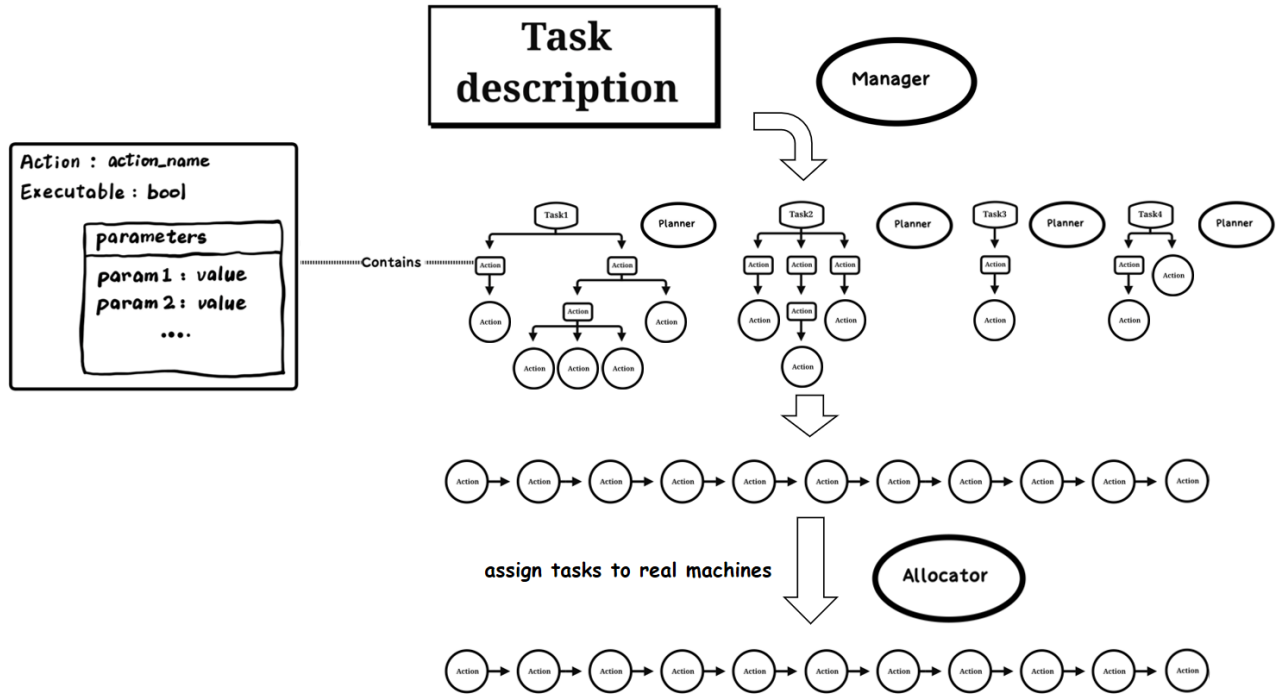

**Fig. 2.** Schematic of the three types of entities cooperating to generate executable task sequences, and a diagram of the basic content required for a single node in the task tree.

From the previous process of task tree, forest, and executable task sequence generation, we can abstract three types of entities. Their interactions are as follows:

1. The Manager obtains the task description and generates the total task word sequence and general parameters for all total task words.
2. Find the corresponding Planner for each total task word.
3. Each Planner generates task word sub-sequences and parameters based on the task word and parameters, and further generates the next step's sub-task sequences and parameters according to the generated task word sub-sequences and parameters, until all the task words in the leaves of the final tree are executable task words, obtaining a complete task tree.
4. The Allocator assigns the operations to be executed by the actual machines based on the task tree and task description, and assigns specific machines to the nodes of the task tree.

The entities design described is as table 2.

**Table. 2.** Explanation of the three types of entities required in the process of decoupling and coordination.

| Entities Design |
| --- |
| **Manager** — Obtains the task description and generates a flat sequence of overall task words, then invokes the entity to generate the task tree for each overall task word. |
| **Planner** — Generates the task tree for a specific overall task word. |
| **Allocator** — Allocates tasks to specific machines based on the task description and the forest. |

The specific pseudocode logic described is as follows:

**Algorithm 1: Generate Task Tree**

**Input:** A high level instruction i; state information s; manager, object that can generate a base task list based on its map, a set of valid general operations(each can be generated by a planner) and their description(about what they can do, the meaning of their parameters, some possible subtask); planners, a set of objects that can generate different task, and they each have its own map, a set of valid operations and their description(about what they can do and the meaning of their parameters); allocator, an object who can allocate robots to specific node/leaf in the task tree based on i and s.
**Output:** A tree whose leaf represents a function which the robot can execute

```
1: Generate_Task_Tree (i,s,manager,planners,allocator)
2:   base_list=manager.initialize_base_list(i,s)
3:   for task in base_list:
4:     planner=manager.find_planner(task)
5:     If (planner exist)
6:       task=planner.generate(task)
7:     Else
8:       error
9:     End If
10:  base_list = allocator.allocate_robot(base_list)
11:  Return base_list
12: End Procedure
```

**Algorithm 2: planner.generate(action: str, node: TaskNode, general_info: str)**

**Input:** action - the action word for the task
  node - the root node of the task tree
  general_info - general information for the task
**Output:** None (modifies the task tree)

1: Procedure generate(action, node, general_info)
2:    Check if the task word is a key in the map dictionary
3:      If it is a valid task word, continue
4:    End Check
5:
6:    While (the task tree still has leaf nodes with is_func =0)
7:      For each leaf node with is_func = 0
8:        Get the action and parameters of the node
9:        Retrieve the task word from the Planner's mapper
10:       If the task word is not found, throw an exception
11:       Get the rules and action restrictions for the task word
12:       Organize action, parameters, rules, action restrictions, and general_info as a prompt
13:       Send the prompt to the language model
14:       Receive the response in JSON format from the language model, which contains the sequence of overall task words and corresponding parameter strings
15:       Log the response
16:       Convert the JSON string to a dictionary
17:       For each object in the dictionary
18:         Create a new TaskNode based on the "action" and "parameter" values:
19:         Check if the action exists in the Planner's mapper
20:         If it exists, get the is_func value for the action and create a corresponding TaskNode object based on the "action" and "parameter" of the current object
21:         End Check
22:         Add the newly created TaskNode as a child of the leaf node being processed in the current iteration
23:       End For
24:    End While
25: End Procedure

## IV. FEASIBILITY EXPERIMENT DESIGN

First, we manually abstract simple assembly tasks, design available instruction sets, task word sets, and possible sub-task sequences of tasks to simulate the role of professional knowledge and verify the feasibility of its output meeting the required instruction sequence.

We use ChatGPT[28], the most advanced large language model currently from OpenAI, to test the method we propose. We inform the model of the input and output templates, provide five examples from complex to simple, repeat them two to three times, and then input them, asking for the output of task planning results.

The version of ChatGPT we used in the test was released on February 13, 2023. We used both ChatGPT 3.5 and ChatGPT 4, and obtained results through their default online interfaces. Since ChatGPT does not yet provide users with hyperparameter configurations, and its generated results will vary each time, we recorded the results of its three generations for the same input.

### A. Example Design and Test Case Design

We design more complex examples for training to verify the feasibility of the model to understand the output on simpler tasks. The possible sub-task words of the task words have the following generative relationship: For the consideration of verifying the minimal system, only the PlanAssembly task in the instructions has a loop, which may continue to decompose the task according to the recursive conditions until the task scale is small enough and the complexity is low enough to generate the AssembleParts command. We designed a task to assemble a solar-powered car and a task to assemble a toy desk lamp for testing. Detailed instructions and Prompts can be found on our Github.

Since PlanAssembly needs to decompose more complex tasks into several parts and then continue to decompose, we use a description like "perform an action on object A and object B" as an atomic assembly task to measure the complexity of a given task. We wrote some ideal generations for these two tasks and used five examples from complex to simple to provide the model with context information.

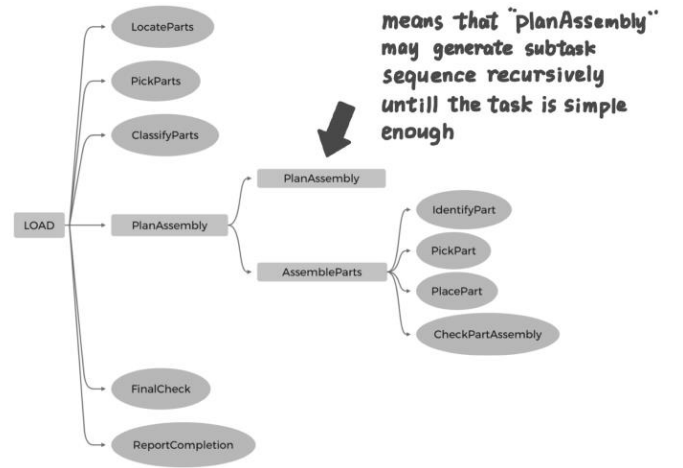

**Fig. 3.** Schematic of the generation relationships among the tasks in the test cases.

### B. Evaluation Method

In order to understand the performance of the proposed method, we use the following indicators for evaluation:

1) Format Success Rate (format_success_rate): It measures whether the skills selected by the model correctly comply with

the logical nesting requirements of the instruction set, regardless of whether they are a successful task sequence that can complete the total task at once. It is measured by the ratio of successful formatting times to total generation times.

2) Parameter Success Rate (parameter_success_rate): For example, if a parameter only accepts one object to be recognized, but the LLM fills in two. We record the number of sub-tasks where all parameters are successfully generated, and take the ratio to the total number of generated sub-tasks.

3) Planning Success Rate (plan_success_rate): It measures whether the generated task sequence successfully executes the total task. As many instructions may have multiple valid solutions simultaneously, we ask 3 human evaluators to judge whether the plan generated by the model can complete the instruction to eliminate this error. If 2 evaluators think the plan is effective, it is marked as successful. It is measured by the ratio of successful times to total generation times.

We examined tasks that require recursive hierarchies, disassembly of assembly steps, and extraction of parameters for single assembly actions.

## V. EXPERIMENT RESULTS

**Table. 3.** Experimental results for tasks that require recursive layering.

| MODEL | FORMAT_SUCCESS_RATE | PARAMETER_SUCCESS_RATE | PLAN_SUCCESS_RATE |
|---|---|---|---|
| GPT 4. | 100% | 81.8% | 100% |
| GPT 3.5 | 100% | 100% | 0% |

**Table. 4.** Experimental results for tasks that disassembly step decomposition tasks.

| MODEL | FORMAT_SUCCESS_RATE | PARAMETER_SUCCESS_RATE | PLAN_SUCCESS_RATE |
|---|---|---|---|
| GPT 4. | 100% | 79.2% | 83.3% |
| GPT 3.5 | 100% | 66.6% | 83.3% |

**Table. 5.** Experimental results for tasks that single assembly action parameter extraction tasks.

| MODEL | FORMAT_SUCCESS_RATE | PARAMETER_SUCCESS_RATE | PLAN_SUCCESS_RATE |
|---|---|---|---|
| GPT 4. | 100% | 100% | 100% |
| GPT 3.5 | 100% | 100% | 100% |

## VI. RESULT ANALYSIS

*A. Feasible Parts*

1) Stably generates the specified code format;
2) Understands the relationship between tasks and sub-tasks, possible sub-task sequences, and generates sub-task sequences according to the requirements of possible sub-tasks;
3) Understands the meaning of parameters and extracts parameters from text descriptions;
4) Understands the logic of recursive tasks and follows the task decomposition structure.

*B. Existing Problems*

1) The logic complexity of processing tasks at a single time is limited;
2) There is ambiguity in the grasp of the number of parts and the precise location of assembly;
3) Errors in parameter extraction occur when the concept of "parts" is unclear, which generally happens when there are other references to this part in the task description. For example, "metal rod" and "drive shaft" refer to the same object in context;
4) It should be noted that the language comprehension ability of GPT3.5 seems not to support it to understand a requirement to recursively decompose tasks. We tried to change its behavior in the interaction, but it was completely unsuccessful.

*C. Possible Solutions*

In response to the problems encountered, we have attempted to propose some possible solutions:
1) Improve the precision of task description;
2) Enhance the precision of thinking structure:
   a. Precise description of the task;
   b. Increase the depth of task decomposition and control the amount of tasks handled at one time.

Adjusting in these aspects can have some effects. Thanks to the convenience that the LLM can change its behavior based on text prompts, just by adding constraint instructions in the task description and providing it with more Prompts, the generation results can be improved to a certain extent.

## VII. CONLUSION

By integrating professional knowledge into the language model, we can enable it to solve professional tasks. This will allow the language model to understand and generate task planning schemes related to professional fields, providing guidance and decision support for robots or other intelligent systems. For example, in the medical field, the language model can generate reasonable diagnosis and treatment plans based on medical knowledge and guidelines; in the financial field, the language model can generate optimized investment portfolio planning based on market analysis and investment strategies. Combining the language model with the thinking framework of professional fields can not only improve the accuracy and efficiency of task planning, but also reduce the involvement of human experts. The language model can become a powerful intelligent assistant, providing real-time decision support and task planning suggestions for professionals.

At present, the method we proposed is still in its infancy. The next step is to continue to optimize its expressive ability, add descriptions of concepts and precise mapping mechanisms, and consider how to better represent the possible subsequence weights. We hope our ideas can inspire further exploration and development.

REFERENCES

[1] Caelan Reed Garrett, Rohan Chitnis, Rachel Holladay, Beomjoon Kim, Tom Silver, Leslie Pack Kaelbling, and Tomás Lozano-Pérez. Integrated task and motion planning. Annual review of control, robotics, and autonomous systems, 4:265–293, 2021

[2] Leslie Pack Kaelbling and Tomás Lozano-Pérez. Hierarchical task and motion planning in the now. In 2011 IEEE International Conference on Robotics and Automation, pages 1470–1477, 2011. doi: 10.1109/ICRA.2011.5980391.

[3] Caelan Reed Garrett, Tomás Lozano-Pérez, and Leslie Pack Kaelbling. Pddlstream: Integrating symbolic planners and blackbox samplers via optimistic adaptive planning. In Proceedings of the International Conference on Automated Planning and Scheduling, volume 30, pages 440–448, 2020.

[4] Danny Driess, Ozgur Oguz, and Marc Toussaint. Hierarchical task and motion planning using logic-geometric programming (hlgp). In RSS Workshop on Robust Task and Motion Planning, 2019.

[5] Zi Wang, Caelan Reed Garrett, Leslie Pack Kaelbling, and Tomás Lozano-Pérez. Active model learning and diverse action sampling for task and motion planning. In 2018 IEEE/RSJ International Conference on Intelligent Robots and Systems (IROS), pages 4107–4114. IEEE, 2018.

[6] Danny Driess, Jung-Su Ha, and Marc Toussaint. Deep visual reasoning: Learning to predict action sequences for task and motion planning from an initial scene image. arXiv preprint arXiv:2006.05398, 2020.

[7] Danny Driess, Ozgur Oguz, Jung-Su Ha, and Marc Toussaint. Deep visual heuristics: Learning feasibility of mixed-integer programs for manipulation planning. In 2020 IEEE International Conference on Robotics and Automation (ICRA), pages 9563–9569. IEEE, 2020.

[8] Danny Driess, Jung-Su Ha, Russ Tedrake, and Marc Toussaint. Learning geometric reasoning and control for long-horizon tasks from visual input. In 2021 IEEE International Conference on Robotics and Automation (ICRA), pages 14298–14305. IEEE, 2021.

[9] Danny Driess, Jung-Su Ha, Russ Tedrake, and Marc Toussaint. Learning geometric reasoning and control for long-horizon tasks from visual input. In 2021 IEEE International Conference on Robotics and Automation (ICRA), pages 14298–14305. IEEE, 2021.

[10] Tom Silver, Ashay Athalye, Joshua B. Tenenbaum, Tomás Lozano-Pérez, and Leslie Pack Kaelbling. Learning neuro-symbolic skills for bilevel planning. In 6th Annual Conference on Robot Learning, 2022. URL https://openreview.net/forum?id=OIaJRUo5UXy.

[11] Tom Silver, Rohan Chitnis, Aidan Curtis, Joshua B Tenenbaum, Tomas Lozano-Perez, and Leslie Pack Kaelbling. Planning with learned object importance in large problem instances using graph neural networks. In Proceedings of the AAAI conference on artificial intelligence, volume 35, pages 11962–11971, 2021.

[12] Rohan Chitnis, Tom Silver, Beomjoon Kim, Leslie Kaelbling, and Tomas Lozano-Perez. Camps: Learning context-specific abstractions for efficient planning in factored mdps. In Conference on Robot Learning, pages 64–79. PMLR, 2021.

[13] Rohan Chitnis, Tom Silver, Joshua B Tenenbaum, Tomas Lozano-Perez, and Leslie Pack Kaelbling. Learning neuro-symbolic relational transition models for bilevel planning. In 2022 IEEE/RSJ International Conference on Intelligent Robots and Systems (IROS), pages 4166–4173. IEEE, 2022.

[14] Aidan Curtis, Tom Silver, Joshua B Tenenbaum, Tomás Lozano-Pérez, and Leslie Kaelbling. Discovering state and action abstractions for generalized task and motion planning. In Proceedings of the AAAI Conference on Artificial Intelligence, volume 36, pages 5377–5384, 2022.

[15] Lin Shao, Toki Migimatsu, Qiang Zhang, Karen Yang, and Jeannette Bohg. Concept2robot: Learning manipulation concepts from instructions and human demonstrations. The International Journal of Robotics Research, 40(12-14):1419–1434, 2021.

[16] Mohit Shridhar, Lucas Manuelli, and Dieter Fox. Cliport: What and where pathways for robotic manipulation. In Conference on Robot Learning, pages 894–906. PMLR, 2022.

[17] Mohit Shridhar, Lucas Manuelli, and Dieter Fox. Perceiver-actor: A multi-task transformer for robotic manipulation. arXiv preprint arXiv:2209.05451, 2022.

[18] Oier Mees, Lukas Hermann, Erick Rosete-Beas, and Wolfram Burgard. Calvin: A benchmark for languageconditioned policy learning for long-horizon robot manipulation tasks. IEEE Robotics and Automation Letters (RA-L), 7(3):7327–7334, 2022.

[19] Anthony Brohan, Noah Brown, Justice Carbajal, Yevgen Chebotar, Joseph Dabis, Chelsea Finn, Keerthana Gopalakrishnan, Karol Hausman, Alex Herzog, Jasmine Hsu, Julian Ibarz, Brian Ichter, Alex Irpan, Tomas Jackson, Sally Jesmonth, Nikhil Joshi, Ryan Julian, Dmitry Kalashnikov, Yuheng Kuang, Isabel Leal, KuangHuei Lee, Sergey Levine, Yao Lu, Utsav Malla, Deeksha Manjunath, Igor Mordatch, Ofir Nachum, Carolina Parada, Jodilyn Peralta, Emily Perez, Karl Pertsch, Jornell Quiambao, Kanishka Rao, Michael Ryoo, Grecia Salazar, Pannag Sanketi, Kevin Sayed, Jaspiar Singh, Sumedh Sontakke, Austin Stone, Clayton Tan, Huong Tran, Vincent Vanhoucke, Steve Vega, Quan Vuong, Fei Xia, Ted Xiao, Peng Xu, Sichun Xu, Tianhe Yu, and Brianna Zitkovich. Rt-1: Robotics transformer for real-world control at scale. In arXiv preprint arXiv:2212.06817, 2022.

[20] Wenlong Huang, Pieter Abbeel, Deepak Pathak, and Igor Mordatch. Language models as zero-shot planners: Extracting actionable knowledge for embodied agents. arXiv preprint arXiv:2201.07207, 2022.

[21] Ishika Singh, Valts Blukis, Arsalan Mousavian, Ankit Goyal, Danfei Xu, Jonathan Tremblay, Dieter Fox, Jesse Thomason, and Animesh Garg. Progprompt: Generating situated robot task plans using large language models. arXiv preprint arXiv:2209.11302, 2022.

[22] M. Ahn, A. Brohan, N. Brown, Y. Chebotar, O. Cortes, B. David, C. Finn, K. Gopalakrishnan, K. Hausman, A. Herzog, et al., Do as i can, not as i say: Grounding language in robotic affordances, arXiv preprint arXiv:2204.01691 (2022).

[23] I. Singh, V. Blukis, A. Mousavian, A. Goyal, D. Xu, J. Tremblay, D. Fox, J. Thomason, A. Garg, Progprompt: Generating situated robot task plans using large language models, arXiv preprint arXiv:2209.11302 (2022).

[24] Cao Y, Lee C S. Robot Behavior-Tree-Based Task Generation with Large Language Models[J]. arXiv preprint arXiv:2302.12927, 2023.

[25] Lin K, Agia C, Migimatsu T, et al. Text2motion: From natural language instructions to feasible plans[J]. arXiv preprint arXiv:2303.12153, 2023.

[26] Christopher Agia, Toki Migimatsu, Jiajun Wu, and Jeannette Bohg. Stap: Sequencing task-agnostic policies. arXiv preprint arXiv:2210.12250, 2022.

[27] OpenAI, How do text-davinci-002 and text-davinci-003 differ?, 2022. URL: https://help.openai.com/en/articles/6779149-how-do-text-davinci-002-and-text-davinci-003-differ.

[28] OpenAI, ChatGPT: Optimizing language models for dialogue, 2022. URL: https://openai. com/blog/chatgpt/.